\documentclass{article}

\usepackage[preprint]{neurips_2024}

\usepackage[utf8]{inputenc} 
\usepackage[T1]{fontenc}    
\usepackage{hyperref}       
\usepackage{url}            
\usepackage{booktabs}       
\usepackage{amsfonts}       
\usepackage{nicefrac}       
\usepackage{microtype}      
\usepackage[table]{xcolor}         
\usepackage{graphicx}
\usepackage{multicol}
\usepackage{multirow}
\usepackage{makecell}
\usepackage{listings}
\usepackage{bm}
\usepackage{subfig}
\usepackage{amsmath}
\usepackage{amsfonts,amssymb}
\usepackage{wrapfig}
\usepackage{appendix}
\usepackage{colortbl}
\usepackage{adjustbox}
\usepackage[table]{xcolor} 

\NewDocumentCommand{\xt}{ mO{} }{\textcolor{red}{\textsuperscript{\textit{XT}}\textsf{\textbf{\small[#1]}}}}

\NewDocumentCommand{\xf}{ mO{} }{\textcolor{blue}{\textsuperscript{\textit{XF}}\textsf{\textbf{\small[#1]}}}}

\title{InfiMM-WebMath-40B: Advancing Multimodal Pre-Training for Enhanced Mathematical Reasoning}

\author{%
Xiaotian Han$^{1}$\thanks{Equal contributions.} \quad Yiren Jian$^{1*}$ \quad Xuefeng Hu$^{1*}$ \quad Haogeng Liu$^{1,2,*}$ \quad Yiqi Wang$^{1*}$ \\
\textbf{Qihang Fan}$^{1,2}$ \quad \textbf{Yuang Ai}$^{1,2}$ \quad \textbf{Huaibo Huang}$^2$ \quad \textbf{Ran He}$^2$ \quad \textbf{Zhenheng Yang}$^1$ \\ 
\textbf{Quanzeng You}$^{1}$\thanks{Corresponding author.}\\
$^1$ByteDance, Inc \quad $^2$Chinese Academy of Sciences\\
}

\begin{document}

\maketitle

\begin{abstract}
Pre-training on large-scale, high-quality datasets is crucial for enhancing the reasoning capabilities of Large Language Models (LLMs), especially in specialized domains such as mathematics.
Despite the recognized importance, the Multimodal LLMs (MLLMs) field currently lacks a comprehensive open-source pre-training dataset specifically designed for mathematical reasoning. 
To address this gap, we introduce InfiMM-WebMath-40B, a high-quality dataset of interleaved image-text documents. 
It comprises 24 million web pages, 85 million associated image URLs, and 40 billion text tokens, all meticulously extracted and filtered from CommonCrawl.
We provide a detailed overview of our data collection and processing pipeline.
To demonstrate the robustness of InfiMM-WebMath-40B, we conducted evaluations in both text-only and multimodal settings. 
Our evaluations on text-only benchmarks show that, despite utilizing only 40 billion tokens, our dataset significantly enhances the performance of our 1.3B model, delivering results comparable to DeepSeekMath-1.3B, which uses 120 billion tokens for the same model size. 
Nevertheless, with the introduction of our multi-modal math pre-training dataset, our models set a new state-of-the-art among open-source models on multi-modal math benchmarks such as MathVerse and We-Math. 
We release our data at \textcolor{red}{\href{https://huggingface.co/datasets/Infi-MM/InfiMM-WebMath-40B}{https://huggingface.co/datasets/Infi-MM/InfiMM-WebMath-40B}}.
\end{abstract}

\section{Introduction}
Recently, Large Language Models (LLMs) have exhibited significant advancements, enhancing their capabilities in complex reasoning and multi-step mathematical problem-solving. 
This progress is driven by larger training scales, innovative inference techniques like Chain-of-Thought (CoT) prompting~\cite{wei2022chain}, and the rich diversity of training data. 
These developments have significantly advanced both proprietary models like GPT-4o~\cite{gpt4o}, Claude 3.5 Sonnet~\cite{claude}, and open-source variants such as Llama 3.1~\cite{dubey2024llama}. 
Notably, these models demonstrate enhanced understanding and reasoning, evident in their ability to tackle challenges from basic grade-school word problems in GSM8K~\cite{cobbe2021training} to high school competition-level tasks in MATH~\cite{hendrycks2021measuring}.

In parallel to general-purpose models, specialized smaller LLMs have achieved significant progress. 
Models such as DeepSeekMath-7B~\cite{shao2024deepseekmath} and InternLM-Math~\cite{ying2024internlm} have specialized capabilities in mathematics, emphasizing their effectiveness within focused domains. 
Furthermore, LLM-based formal mathematics proving systems such as Alpha-Proof~\cite{alphaproof} and DeepSeek-Prover~\cite{xin2024deepseek} have shown notable success in solving complex International Mathematics Olympiad (IMO)-level problems. 
These achievements further highlight the potential of LLMs, both large and small, to tackle sophisticated reasoning tasks that approximate human-level capabilities.

While most mathematical knowledge is typically encoded in language, visual elements like figures, diagrams, and geometric plots are pivotal for intuitively understanding abstract mathematical concepts.
Recognizing this, several Multimodal Large Language Models (MLLMs) such as G-LLaVA~\cite{gao2023g}, Math-LLaVA~\cite{shi2024math}, LLaVA-Next~\cite{liu2024llava}, and MAVIS~\cite{zhang2024mavis} 
have been developed to enhance reasoning capabilities by integrating these multimodal inputs. 
These models integrate vision modalities using visual embeddings from pre-trained models such as CLIP~\cite{radford2021learning} and SigLIP~\cite{zhai2023sigmoid}, enhancing their ability to process and reason with visually represented mathematical concepts.
They refine their capabilities through multi-modal instruction tuning with specialized datasets focused on mathematical instruction, such as Geo170k~\cite{gao2023gllavasolvinggeometricproblem,cai2024geogpt4vgeometricmultimodallarge}, MathV360K~\cite{shi2024mathllavabootstrappingmathematicalreasoning}, and MAVIS-Instruct~\cite{zhang2024mavismathematicalvisualinstruction}.

While many of these models show promising progress, studies suggest that introducing new knowledge during the instruction fine-tuning stage can be challenging~\cite{zhu2023physics}, often increasing the risk of hallucinations~\cite{gekhman2024does}, particularly due to limitations in data scale, format, and training strategies. In contrast, large corporations often have access to high-quality, large-scale proprietary pre-training datasets, giving them an advantage in developing MLLMs with strong mathematical reasoning capabilities. However, despite growing demand, publicly available large-scale pre-training datasets that integrate both textual and visual mathematical data remain scarce, significantly hindering progress within the open-source community, which depends on accessible data to drive research and develop truly effective MLLMs.

In this work, we introduce \textbf{InfiMM-WebMath-40B}, the first publicly available large-scale multimodal mathematics pre-training dataset. 
This dataset marks a significant milestone for the open-source community, addressing the long-standing gap in publicly available multimodal math data.
InfiMM-WebMath-40B comprises 24 million mathematics and science-related web documents, including 85 million image URLs and approximately 40 billion text tokens, offering an unprecedented resource for training and fine-tuning Multimodal Large Language Models (MLLMs).

Inspired by previous efforts like OpenWebMath~\cite{paster2023openwebmath} and OBELICS~\cite{laurençon2023obelicsopenwebscalefiltered}, we construct our dataset by filtering mathematical and scientific content from the CommonCrawl~\cite{commoncrawl} repository, preserving full multimodal web documents with interleaved images and text. 
We start with 44 snapshots of CommonCrawl data from 2019 to 2023. 
After filtering for Chinese and English webpages, we are left with 57.2 billion web documents. 
Through a rigorous series of filtering processes—beginning with model-based language filtering and deduplication—we apply both rule-based and model-based filtering using fastText~\cite{joulin2016fasttext}, which narrows the collection down to 24 million high-quality mathematics and science-related web documents.

To showcase the potential of InfiMM-WebMath-40B, we conduct extensive experiments using recent benchmarks such as MathVerse~\cite{zhang2024mathversedoesmultimodalllm} and WeMath~\cite{qiao2024wemathdoeslargemultimodal}. 
Our preliminary results demonstrate the effectiveness of the dataset in enhancing multimodal mathematical reasoning, further validating its importance for both open-source research and the broader AI community.

Our contributions in this work are as follows:
\begin{itemize}
    \item First, we introduce InfiMM-WebMath-40B, the first publicly available, large-scale multimodal dataset specifically designed for mathematical pre-training. 
    This dataset significantly contributes to filling a critical gap in the open-source community by providing a vast collection of high-quality text and image data to advance the mathematical reasoning capabilities of Multimodal Large Language Models (MLLMs).
    \item Second, we detail the comprehensive preprocessing pipeline used to filter and align mathematics and science-related content from CommonCrawl, ensuring the quality and relevance of the dataset.
    \item Lastly, we evaluate the impact of continuing pre-training on  InfiMM-WebMath-40B through extensive experiments.
    Our InfiMM-Math models, trained on top of this dataset, demonstrate strong performance on mathematical reasoning benchmarks, particularly for complex multimodal problems. 
    We believe future research can further unlock MLLM mathematical reasoning capabilities based on our InfiMM-WebMath-40B dataset.
\end{itemize}
\section{Related Work}

\subsection{LLM for Math}

The use of LLMs for mathematical reasoning has been explored in several studies. GPT-3~\cite{brown2020languagemodelsfewshotlearners} already demonstrated the ability to solve basic arithmetic and algebraic problems. 
However, it was observed that the model may produce incorrect or misleading explanations. 
To better assess the progress of LLMs in mathematical reasoning and encourage their improvement, several math-specific evaluation benchmarks and training datasets have been introduced.

\textbf{Evaluation Benchmarks}\quad The GSM8K~\cite{cobbe2021trainingverifierssolvemath} dataset focuses on grad school-level math problems, highlighting models' ability in basic arithmetic and reasoning. 
In contrast, the MATH dataset~\cite{hendrycksmath2021} targets more advanced topics, covering high school and undergraduate-level mathematics.
MMLU~\cite{hendrycks2021measuringmassivemultitasklanguage} assesses LLM performance across multiple subjects, with its STEM section providing insights into the models' capabilities in scientific and mathematical reasoning. 
Additionally, benchmarks such as SAT~\cite{azerbayevllemma} and OCW~\cite{naeini2023largelanguagemodelsfixated} evaluate LLMs on standardized test problems, further probing their mathematical capabilities. 
Recent research has also introduced specialized benchmarks for evaluating logical and counterfactual reasoning in mathematical contexts, such as MalAlgoQA~\cite{liu2024malalgoqapedagogicalapproachevaluating} and MathCheck-GSM~\cite{zhou2024modelreallygoodmath}.
Together, these benchmarks offer a comprehensive framework for assessing the strengths and limitations of LLMs in mathematical reasoning.

\textbf{Training Datasets}\quad Several large-scale math-specific pre-training datasets have been introduced to enhance the mathematical reasoning capabilities of LLMs.
Notable proprietary datasets include WebMath~\cite{polu2020generativelanguagemodelingautomated}, developed by OpenAI, which comprises 35 billion tokens from Github, arXiv, and Math StackExchange.
MathMix~\cite{lightman2023letsverifystepstep}, containing 1 billion high-quality mathematical tokens, has been used to fine-tune GPT-4.
Math Web Pages~\cite{lewkowycz2022solvingquantitativereasoningproblems} contributes 17.5 billion LaTex-based tokens for training models like Minerva. 
Similarly, the DeepSeekMath Corpus~\cite{shao2024deepseekmathpushinglimitsmathematical} offers 120 billion tokens used in training the DeepSeekMath model.
Among the open-source datasets, AMPS~\cite{hendrycks2021measuring} includes over 100,000 Khan Academy problems and 5 million problems generated using Mathematica.
NaturalProofs~\cite{welleck2021naturalproofs} offers 32,000 theorem statements and proofs, along with 14,000 definitions sourced from ProofWiki and other platforms.
OpenWebMath~\cite{paster2023openwebmathopendatasethighquality} filters 14.7 billion tokens of mathematical content from CommonCrawl. MathPile~\cite{wang2023mathpile}, which aggregates data from textbooks, arXiv, Wikipedia, ProofWiki, StackExchange, and other web pages, contains 9.5B tokens.
Proof-Pile-2~\cite{azerbayev2023llemma} combines several sources, including OpenWebMath~\cite{paster2023openwebmathopendatasethighquality}, AlgebraicStack (10.3B tokens of mathematical code), and arXiv papers (28.0B tokens)~\cite{together2023redpajama}. MathInstruct~\cite{yue2023mammothbuildingmathgeneralist} compiles data from 13 math datasets, incorporating intermediate reasoning steps to facilitate supervised fine-tuning of LLMs. 

These datasets, both proprietary and open-source, play a critical role in advancing the mathematical reasoning capabilities of LLMs, enabling more robust and accurate problem-solving across a wide range of mathematical domains.

\subsection{MLLM for Math}

The rapid evolution of MLLMs has sparked significant interest in enhancing their capabilities for multi-modal reasoning~\cite{wang2024exploringreasoningabilitiesmultimodal}. 
This section outlines the key benchmarks and training datasets developed to evaluate and improve MLLMs' mathematical reasoning abilities.

\textbf{Evaluation Benchmarks}\quad
Benchmarks such as GeoEval~\cite{zhang2024geoevalbenchmarkevaluatingllms}, Geometry3K~\cite{lu2021intergpsinterpretablegeometryproblem}, and GeomVerse~\cite{kazemi2023geomversesystematicevaluationlarge} focus on evaluating MLLM performance in plane geometry problems.
While ChartX~\cite{xia2024chartxchartvlmversatile}, ChartQA~\cite{masry2022chartqabenchmarkquestionanswering}, and ChartBench~\cite{xu2024chartbenchbenchmarkcomplexvisual} target the understanding and reasoning of charts and graphical plots. 
Comprehensive math benchmarks like MathVista~\cite{lu2024mathvistaevaluatingmathematicalreasoning}, MathVerse~\cite{zhang2024mathversedoesmultimodalllm}, MathVision~\cite{wang2024measuringmultimodalmathematicalreasoning}, and We-Math~\cite{qiao2024wemathdoeslargemultimodal} combine tasks across plane geometry, solid geometry, and chart interpretation. 
These benchmarks provide a robust framework for assessing the strengths and limitations of MLLMs in mathematical reasoning.

\textbf{Training Datasets}\quad The development of multimodal instruction fine-tuning datasets has been crucial in advancing MLLMs' capabilities for mathematical reasoning.
Key contributions include GeoGPT4V~\cite{cai2024geogpt4vgeometricmultimodallarge}, Geo170k~\cite{gao2023gllavasolvinggeometricproblem}, MathV360K~\cite{shi2024mathllavabootstrappingmathematicalreasoning}, and MAVIS~\cite{zhang2024mavismathematicalvisualinstruction}. 
Most of these instruction fine-tuning datasets are synthesized by ChatGPT using existing public datasets. 
While large-scale open source interleaved multi-modal pre-training datasets, such as  Multimodal-C4~\cite{zhu2023multimodalc4openbillionscale} and OBLICS~\cite{laurençon2023obelicsopenwebscalefiltered}, have been constructed, these datasets are not specifically designed for MLLM mathematical reasoning. 
MINT-1T~\cite{awadalla2024mint1tscalingopensourcemultimodal}, which includes parsed PDFs and arXiv data, aims to enhance MLLMs' reasoning capability.
Multimodal ArXiv~\cite{li2024multimodalarxivdatasetimproving} builds high-quality captioning and instruction tuning datasets from arXiv papers, providing further opportunities for improving reasoning in mathematical contexts. 

The aforementioned datasets have been proposed for instruction tuning of MLLMs to tackle mathematical reasoning problems, leading to the development of several advanced math-focused MLLMs. 
However, despite these advances, a critical gap remains in the community's efforts: the lack of high-quality, large-scale, multimodal pre-training datasets specifically designed for MLLMs. 
In this work, we address this gap by constructing and publicly releasing a high-quality, multimodal, interleaved mathematical dataset specifically designed for MLLM pre-training.
This resource aims to catalyze further progress in the mathematical reasoning capabilities of MLLMs by providing a more robust foundation for pre-training.
\section{Dataset Construction}\label{sec:data-collection}
In this section, we outline the methodology employed in constructing the InfiMM-WebMath-40B dataset.  
Our dataset comprises approximately 24 million webpages, encompassing 40 billion text tokens and 85 million image URLs.
The objective of this study is to develop a large-scale multimodal math dataset that integrates interleaved image and text data.

To develop our dataset, we source content from the CommonCrawl archives, adhering to methodologies established by previous large-scale data constructions for pre-training language models as discussed in works such as RefinedWeb~\cite{penedo2023refinedweb}, DataComp~\cite{li2024datacomp} and further refined by FineWeb~\cite{penedo2024finewebdatasetsdecantingweb}.
Building upon the methodology established in the construction of the OBELICS dataset \cite{laurençon2023obelicsopenwebscalefiltered}, we enhance our dataset with interleaved text and corresponding image URLs. 
The final stage involves the parallel downloading of images based on these URLs. 

\subsection{Text-only Data Curation Pipeline}
\label{sec:text-only-data-collection}

\begin{figure}[h]
  \centering
  \includegraphics[width=0.9\linewidth]{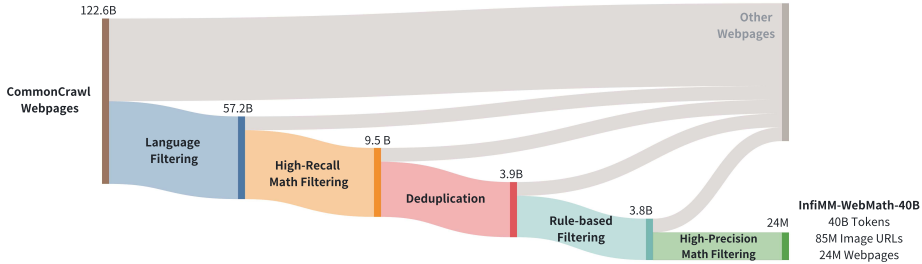}
  \caption{InfiMM-WebMath-40B data curation pipeline.}
  \label{fig:tongji}
\end{figure}

\subsubsection{Text Extraction and Language Filtering}
After careful investigation, we chose Trafilatura~\footnote{\url{https://trafilatura.readthedocs.io/en/latest/}} as the library for extracting text from web pages, as it is a widely adopted Python library for web scraping . 
This library has also been used extensively in the construction of other open-source, text-only datasets, such as RefinedWeb~\cite{penedo2023refinedweb} and FineWeb~\cite{penedo2024finewebdatasetsdecantingweb}. 
However, Trafilatura's design focuses on plain text, omitting math-related equations and symbols—crucial elements for our dataset. 
Consequently, the subsequent section will outline our development of a specialized extraction tool tailored for math-related content.
We initially use Trafilatura to streamline the data extraction process, substantially reducing the volume of data for further processing in our data pipeline.

Following the methodology used in DeepSeekMath~\cite{shao2024deepseekmath}, we focus on retaining only Chinese and English content when constructing our dataset. 
To achieve this, we apply language filtering to the CommonCrawl repositories, which originally comprise approximately 122 billion webpages, as shown in \figureautorefname~\ref{fig:tongji}.

For language detection, we employ a fastText language identification model~\cite{joulin2016bag}, which offers an effective balance between speed and accuracy, making it well-suited for the initial filtering phase. 
This language filtering process significantly reduces the dataset size, lowering the number of pages from 122 billion to 57.2 billion.

\subsubsection{Mathematical Content Extraction}
\label{sec:mathextraction}
Extracting text from HTML in the field of mathematics presents unique challenges, particularly due to the limited availability of specialized tools that can accurately handle mathematical content, such as LaTeX equations. 
Most standard extraction tools are not equipped to process this specialized content, resulting in gaps in accurate content retrieval.
Additionally, capturing image URLs and their corresponding positions in the text is equally important, as both are critical for constructing our dataset comprehensively.

To address these issues, we evaluate several open-source HTML extraction tools to balance high-quality text extraction, accurate math content retrieval, and precise extraction of image URLs. 
Ultimately, we choose to build on top of Resiliparse~\footnote{\url{https://github.com/chatnoir-eu/chatnoir-resiliparse}}, leveraging its extensibility and robust data extraction capabilities as a foundation for our development.

\begin{figure}[h]
  \centering
  \includegraphics[width=\linewidth]{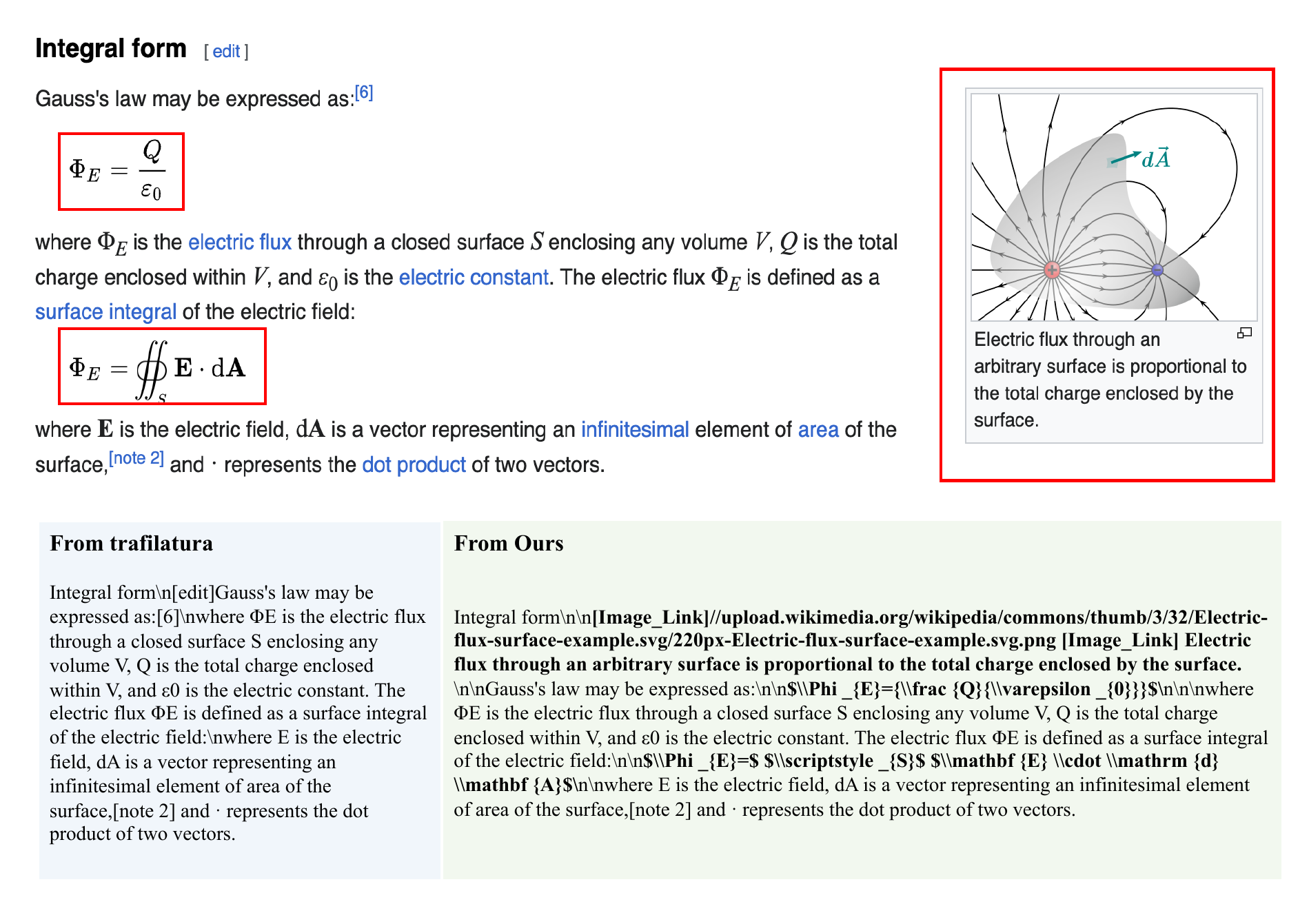}
  \caption{A comparative illustration of extraction results from a Wikipedia webpage using Trafilatura and our enhanced version of Resiliparse, highlighting the successful retrieval of mathematical equations and image URLs.}
  \label{fig:text_extraction}
\end{figure}

\figureautorefname~\ref{fig:text_extraction} illustrates a comparison of extraction results between Trafilatura and our enhanced version of Resiliparse.
Our tool successfully extracts both the mathematical equations and image URLs, as highlighted in the red boxes in the screenshot from a Wikipedia webpage.

\subsubsection{High-Recall Filtering for Mathematical Content}
\label{sec:initial_fasttext}
Building upon the math content extracted previously, we have developed an initial filtering methodology inspired by DeepSeek-Math~\cite{shao2024deepseekmath}. 
We begin by randomly sampling half a million data points from the OpenWebMath~\cite{paster2023openwebmath} dataset, complementing these with an equal number of negative samples from our earlier extracted math content. 
This creates a well-balanced training set, which is crucial for effective classifier development. 
Notably, we removed image URL links from the negative samples to standardize the data format across both positive and negative samples.

Leveraging this balanced dataset, we trained a fastText classifier specifically designed to filter mathematical content. 
This model significantly streamlined our dataset, reducing the volume from 57.2 billion samples to 9.5 billion. 
We set a probability threshold of 0.4 to prioritize recall over precision, allowing for a more inclusive filtering process. 
This lower threshold ensures that we capture a broader range of mathematical content, providing a foundation for subsequent refinement while minimizing the risk of excluding relevant data too early in our data curation pipeline.

\subsubsection{Deduplication}
Removing duplicate webpages from training data is known to positively impact model performance~\cite{lee2021deduplicating}. 
To address this, we implement comprehensive deduplication strategies, including both content and URL deduplication, to eliminate redundant content.

\textbf{Content Deduplication}\quad
We employ MinHash~\cite{broder1997resemblance}, a fuzzy hashing technique widely recognized as a standard in data deduplication~\cite{penedo2023refinedweb, penedo2024finewebdatasetsdecantingweb}.
Following the FineWeb methodology \cite{penedo2024finewebdatasetsdecantingweb}, we apply MinHash-based deduplication within each individual snapshot instead of global deduplication across all data snapshots, as suggested by experimental results. This approach significantly reduces computational demands and data volume.
Subsequently, we extend our deduplication to encompass neighboring snapshot pairs, aiming to further compress the dataset size.
This two-step process is projected to decrease the total dataset volume by approximately 43\%, which equates to a substantial reduction in absolute data size from 9.5 billion to 5.4 billion samples.

\textbf{URL Deduplication}\quad 
Following content deduplication, we apply an exact matching technique for URL deduplication, which further reduces the sample volume from 5.4 billion to 3.9 billion. 
This process is conducted within snapshots from the same year, balancing resource efficiency with data quality.
When identical URLs are identified, we selectively retain content from the more recent snapshot, ensuring that the dataset not only eliminates duplicates but also maintains the most current and relevant information. 

\subsubsection{Rule-based Filtering}
For rule-based filtering methods, we initially explore several filtering rules used in the C4~\cite{2019t5} and OBELICS~\cite{laurençon2023obelicsopenwebscalefiltered} datasets.
However, we discover that most of these rules are not well-suited for a mathematical corpus, as they inadvertently filter out valuable content. 
Ultimately, we retain only the most essential rules to preserve the integrity of the mathematical data. The rules we apply include:

\begin{itemize}
\item{\textbf{Remove Lorem Ipsum}}\quad We remove all short-length documents containing ``lorem ipsum'' while preserving longer documents that may still contain useful content.
\item{\textbf{Punctuation Ratio}}\quad The ``punctuation ratio'' rule proved unsuitable for the Chinese corpus, so we restricted its use to English documents, removing only those with a punctuation ratio exceeding 0.3.
\item{\textbf{NSFW Content}}\quad We filtered out any documents containing NSFW words to ensure the dataset’s appropriateness for academic and research use.
\item{\textbf{Unicode Errors}}\quad Documents containing Unicode errors were excluded to maintain the overall quality and usability of the dataset.
\end{itemize}

Due to the thorough data filtering applied in the earlier steps, we end up removing only about 3\% of the samples based on the above rules, which resulted in 3.8 billion remaining samples. 

\subsubsection{High-Precision Filtering for Mathematical Content}
\label{sec:high-precision}
Our initial fastText classifier (Sec.~\ref{sec:initial_fasttext}), employed positive samples from the OpenWebMath dataset~\cite{paster2023openwebmath} and negative samples from our own data extraction from CommonCrawl. 
To enhance the accuracy of our labeling process, we leveraged the LLaMA3-70B-Instruct model~\cite{dubey2024llama}, employing prompt formats inspired by the FineWeb-Edu dataset~\cite{lozhkov2024fineweb-edu}. This allowed us to score the mathematical quality of each sample on a scale from 0 to 10. 
Table~\ref{tab:llama3:prompt} in Appendix~\ref{sec:llama3:prompt} displays the full prompt.

From the data remaining after rule-based filtering, we randomly sample approximately one million entries. 
Using the vLLM inference engine~\cite{kwon2023efficient}, we assign math quality scores and apply a threshold of 6 to select positive samples for training our updated fastText classifier. 
This process yielded 640,000 positive samples.
We then randomly select an equivalent number of 640,000 negative samples as outlined in Sec.~\ref{sec:initial_fasttext}. 
These positive and negative samples are combined to train the new fastText classifier.\footnote{We also employ an LLM-based classifier for high-precision filtering, Appendix~\ref{sec:ablation:hp:filtering} shows the comparison.}

During fastText training, we implement data cleaning rules to optimize the model's performance for mathematical content. 
Mathematical texts pose unique challenges due to specialized terminology, symbols, formulas, and numeric data, which differ from typical natural language and require more refined preprocessing techniques.

Our goal is to standardize and simplify the input training data while preserving essential mathematical information. 
Key considerations include maintaining consistency in token representation, minimizing noise from extraneous characters, and standardizing numeric values. 
The following steps reflect this approach:
\begin{itemize} 
\item Utilizing the SpaCy English language model (\texttt{en\_core\_web\_sm}), we preprocess the input text, tokenize it, and process each token by converting it to its lowercase and lemmatized form. Common placeholders are replaced, certain non-alphanumeric characters are removed, and patterns of special characters like dashes and underscores are normalized. We also strip any unnecessary whitespace, ensuring the text is well-prepared for downstream processing. 
\item All numeric values are replaced with the <NUM> placeholder to standardize the representation, and line breaks along with carriage returns are removed. Tokens exceeding 100 characters in English are discarded.
\end{itemize}

These preprocessing steps notably enhance the classifier’s performance. 
For evaluation, we use all samples in the Geometry3K~\cite{lu2021intergpsinterpretablegeometryproblem} benchmark as positive examples of mathematical content. 
With our refined preprocessing techniques, fastText's accuracy increases from 48.74\% to 72.15\%.

\subsubsection{Evaluation on Text-Only Filtering}
To provide a preliminary evaluation of the quality of our filtered dataset, we continue pretraining a deepseek-coder-1.3b-base model for one epoch using the filtered mathematical content in Sec.~\ref{sec:high-precision}, excluding image URLs.
We validate the effectiveness of our math-related filtering with a few-shot evaluation using the GSM8K~\cite{cobbe2021training} and the STEM sections of the MMLU~\cite{hendrycks2021measuringmassivemultitasklanguage} benchmark.

\begin{table}[ht]
\centering
\caption{Evaluation of models on GSM8K and MMLU (STEM). The baseline is the deepseek-coder-1.3b-base without any training.}
\label{tab:model_eval_gsm8k_mmlu}
\begin{tabular}{lcc}
\toprule
Training Corpus         & GSM8K        & MMLU (STEM) \\
\midrule
Baseline                &  4.8         & 25.6 \\
OpenWebMath~\cite{paster2023openwebmathopendatasethighquality}             & 11.0         & 29.6 \\
DeepSeekMath Corpus~\cite{shao2024deepseekmathpushinglimitsmathematical}     & 23.8         & 33.1 \\
InfiMM-WebMath-40B (text)   & 26.1         & 35.6 \\
\bottomrule
\end{tabular}
\end{table}

As shown in Table~\ref{tab:model_eval_gsm8k_mmlu}, the model trained on our InfiMM-WebMath-40B text-only dataset demonstrates competitive performance compared to OpenWebMath and the DeepSeekMath Corpus, highlighting the high quality of our dataset and the effectiveness of our filtering procedures.

\subsection{Multimodal Data Construction}
In Sec.~\ref{sec:mathextraction}, we successfully extract image URLs from each webpage as part of the multimodal dataset construction process. 
Following the OBELICS format \cite{laurençon2023obelicsopenwebscalefiltered}, we organize the data into two parallel lists—one for text and one for image URLs—to preserve the original sequence of content on the webpages. 
After applying our filtering pipeline in Sec.~\ref{sec:text-only-data-collection}, we obtain 24 million interleaved mathematical documents, consisting of 85 million image URLs and 40 billion text tokens. 
For the release of the InfiMM-WebMath-40B dataset, we provide the interleaved text and image URLs, adhering to the OBELICS dataset format.

Next, we proceed to download the images from the remaining 85 million image URLs to construct our version of the multimodal dataset for experiments in subsequent sections. 
Upon initial inspection, we find that many URLs are duplicates, often linking to the same equations, mathematical diagrams, graphical plots, or common background images. 
To optimize this process, we first implement deduplication and filtering to focus on unique and relevant images.

To avoid downloading duplicates, we retain only unique image URLs. 
Additionally, we remove URLs that appear more than 10 times across samples or originate from documents containing over 100 images, as these are likely to be noisy or irrelevant.
This refinement reduces the total to 23 million unique image URLs. We further apply keyword filtering, keeping only URLs that begin with ``https'' and excluding those containing terms such as ``logo'', ``banner'', ``avatar'', or ``icon'', reducing the set to 22 million URLs.

From this filtered set, we successfully download 14 million unique images\footnote{While additional nodes, IPs, or proxies could improve download success, this dataset version strikes a good balance between resource use and size for our experiments.}. These images are then reintegrated into their original positions within the interleaved image-text documents, ultimately yielding 24 million records with a total of 28 million downloaded images, due to the shared use of images across multiple webpages.

\section{Experiments}
We start by introducing our model architectures in Sec.~\ref{sec:models}, followed by a discussion of continue pre-training and instruction fine-tuning in Sec.~\ref{sec:training-details}. 
Our primary goal is to verify the effectiveness of the collected pre-training data. 
We conduct experiments to evaluate the trained model, using selected architectures and training stages. 
The evaluation results are discussed in Sec.~\ref{sec:eval}.

\subsection{Model Architectures}\label{sec:models}
Our model architecture design aligns with the latest advancements in vision-language learning \cite{liu2024visual, li2023blip}, incorporating a visual encoder followed by a vision-to-language connector and an LLM decoder. Specifically, we employ the SigLip model \texttt{siglip-so400m-patch14-384} to extract visual features. To balance computational efficiency with performance, we use a 3-layer Perceiver Resampler \cite{jaegle2021perceiver} with 64 latents to reduce the number of tokens/features per image to 64. These visual token/feature embeddings are then concatenated with text embeddings before being fed into the LLMs. In this study, we experiment with two different LLMs from DeepSeek-Coder \cite{guo2024deepseek}: \texttt{deepseek-coder-1.3b-base} and \texttt{deepseek-coder-7b-v1.5}.

\subsection{Training Details}\label{sec:training-details}
In this section, we detail the training data and processes involved in our three-stage training approach: modality alignment, continue pre-training using InfiMM-WebMath-40B, and instruction fine-tuning.

\subsubsection{Modality Alignment Stage}
In this stage, we utilize general-purpose image-text pairs to align the visual encoder and the LLM via Perceiver Resampler.
The primary objective is to minimize the domain gap between visual and linguistic modalities.
To achieve this, we sample a 8 million image-text pair subset from the DFN-2B dataset~\cite{fangdata} for the alignment training. 
During this stage, the vision encoder and LLM backbone are frozen, and training is focused on the Perceiver Resampler module.
Training is conducted for one epoch using DeepSpeed Zero2, with the AdamW optimizer, configured with a cosine learning rate scheduler, a maximum learning rate of $1e^{-4}$, betas of $(0.9, 0.95)$, and a weight decay of 0.1.

\subsubsection{Continue Pre-training Stage}
We further continue pre-training our models using the InfiMM-WebMath-40B dataset to enhance the model's mathematical knowledge acquisition in a multi-modal setting.
The training is conducted for one epoch using DeepSpeed Zero2, with the AdamW optimizer, configured with a cosine learning rate scheduler, a maximum learning rate of $5e^{-5}$, betas of $(0.9, 0.95)$, and a weight decay of 0.1. 
The context length for training examples is set to 4096, with a maximum of 32 images per example.
During this stage, the visual encoder remains frozen, and training focuses on learning the Perceiver Resampler module (the visual-language connector) and the LLM.

\subsubsection{Instruction Fine-tuning Stage}
In this stage of training, we fine-tune our models using instruction datasets, including PGPS9K \cite{zhang2023multi}, Geo170k \cite{gao2023gllavasolvinggeometricproblem}, TABMWP \cite{lu2023dynamic}, ScienceQA \cite{lu2022learn}, Vflan \cite{chen2024allava}, VisualWebInstruct, AI2D \cite{kembhavi2016diagram}, ChartQA \cite{masry2022chartqabenchmarkquestionanswering}, DocVQA \cite{mathew2021docvqa}, DVQA \cite{kafle2018dvqa}, GeoQA \cite{chen2021geoqa}, and MAVIS \cite{zhang2024mavismathematicalvisualinstruction}. 
We find that incorporating uni-modal text instruction datasets is crucial for enhancing the models' instruction-following capabilities. 
Therefore, we also include pure text instruction datasets such as Math\cite{li2023camel}, MetaMathQA \cite{yu2024metamath}, DART-Math \cite{tong2024dart}, and NuminaMath \cite{numina_math_7b}.
The objective of this stage is to acclimate the models to the common chat templates used in math VQA settings, thereby enabling them to better utilize the mathematical knowledge acquired in the previous stage.

We freeze the vision encoder and update the parameters of the Perceiver Resampler and LLMs. 
As in the previous stages, training is conducted using DeepSpeed Zero2 for one epoch, with the AdamW optimizer, configured with 2000 warmup steps, a maximum learning rate of $5e^{-6}$, betas of $(0.9, 0.95)$, a weight decay of 0.1, and cosine decay to $5e^{-7}$. 
The batch size is set to one per GPU, and the context length of the training examples is set to 4096. 
We utilize 32 A100-80G GPUs for the 1.3b models and 64 A100-80G GPUs for the 7b models. 
We refer to the resulting model as InfiMM-Math.

\subsection{Evaluations}\label{sec:eval}
In the following sections, we discuss the evaluation of our models on two widely adopted multi-modal math-only benchmarks: MathVerse \cite{zhang2024mathversedoesmultimodalllm} and We-Math \cite{qiao2024wemathdoeslargemultimodal}.

\subsubsection{Evaluations on MathVerse}

\begin{table}[t]
\begin{minipage}{\textwidth}
\begin{center}
\caption{Evaluation of models on MathVerse. Following the official MathVerse recommendation, we report the ``w/o'' scores based on the correctness of final answers.}
\begin{tabular}{cccccccc}
\toprule
Model             & \begin{tabular}[c]{@{}c@{}}Base\\ LLM\end{tabular} & All  & \begin{tabular}[c]{@{}c@{}}Text\\ Dominant\end{tabular} & \begin{tabular}[c]{@{}c@{}}Text\\ Lite\end{tabular} & \begin{tabular}[c]{@{}c@{}}Vision\\ Intense\end{tabular} & \begin{tabular}[c]{@{}c@{}}Vision\\ Dominant\end{tabular} & \begin{tabular}[c]{@{}c@{}}Vision\\ Only\end{tabular} \\
\midrule
Human             & \scriptsize{-} & 64.9 & 71.2 & 70.9 & 61.4 & 68.3 & 66.7 \\
\midrule
\multicolumn{8}{c}{\textit{Proprietary Models}} \\
\midrule
GPT-4V            & \scriptsize{N/A} & 39.4 & 54.7 & 41.4 & 34.9 & 34.4 & 31.6 \\
Gemini-Pro        & \scriptsize{N/A} & 23.5 & 26.3 & 23.5 & 23.0 & 22.3 & 22.2 \\
Qwen-VL-Max       & \scriptsize{N/A} & 25.3 & 30.7 & 26.1 & 24.1 & 24.1 & 21.4 \\
\midrule
\multicolumn{8}{c}{\textit{Open-sourced Models}} \\
\midrule
SPHINX-Plus       & \scriptsize{LLaMA2-13B} & 14.0 & 16.3 & 12.8 & 12.9 & 14.7 & 13.2 \\
G-LLaVA           & \scriptsize{LLaMA2-7B} & 15.7 & 22.2 & 20.4 & 16.5 & 12.7 & 6.6 \\
InternLM-XC2      & \scriptsize{InternLM2-7B} & 16.5 & 22.3 & 17.0 & 15.7 & 16.4 & 11.0 \\
Math-LLaVA        & \scriptsize{Vicuna-13B} & 19.0 & 21.2 & 19.8 & 20.2 & 17.6 & 16.4 \\
ShareGPT4V        & \scriptsize{Vicuna-13B} & 17.4 & 21.8 & 20.6 & 18.6 & 16.2 & 9.7 \\
LLaVA-NeXT        & \scriptsize{LLaMA3-8B} & 19.3 & 24.9 & 20.9 & 20.8 & 16.1 & 13.8 \\
LLaVA-NeXT        & \scriptsize{Qwen-1.5-110B} & 24.5 & 31.7 & 24.1 & 24.0 & 22.1 & 20.7 \\
MAVIS             & \scriptsize{Mammoth2-7B} & 27.5 & 41.4 & 29.1 & 27.4 & 24.9 & 14.6 \\
\midrule
\multicolumn{8}{c}{\textit{Our Models}} \\
\midrule
InfiMM-Math       & \scriptsize{DS-Coder-1.3B} & 26.9 & 37.1 & 30.2 & 29.2 & 24.4 & 13.7 \\
InfiMM-Math       & \scriptsize{DS-Coder-1.5-7B} & 34.5 & 46.7 & 32.4 & 38.1 & 32.4 & 15.8 \\
\bottomrule
\end{tabular}
\label{table:eval-mathverse}
\end{center}
\end{minipage}
\end{table}

We evaluate our models on the \texttt{testmini} set of MathVerse, which contains nearly 4,000 instances. Each problem in MathVerse is constructed with varying levels of multi-modal information, including text-dominant, text-lite, vision-intensive, vision-dominant, and vision-only categories. The questions can be either open-ended or multiple-choice.

We follow the three-stage evaluation pipeline outlined in MathVerse. In the first stage, models generate answers based on the provided prompts and images. In the second stage, GPT-4 is employed to extract answers from the models' outputs. Finally, in the third stage, GPT-4 is used to determine whether the extracted answers match the ground truth.

Following the official MathVerse recommendations, we report the ``w/o'' score (based on the correctness of final answers) using the suggested CoT prompts. The results are shown in Table~\ref{table:eval-mathverse}. Our 7B model outperforms all open-source models, including the 110B LLaVA-NeXT. It also surpasses Gemini-Pro and Qwen-VL-Max, trailing only behind GPT-4V. Our model excels in the Text-Dominant, Text-Lite, Vision-Intense, and Vision-Dominant sections, demonstrating its strong multi-modal understanding when processing both text and visual inputs. Our model underperforms in the Vision-Only section, likely due to the limitations of the vision encoder we employed, which processes input images at $384 \times 384$ resolution, whereas LLaVA-NeXT supports $336 \times [(2,2), (1,2), (2,1), (1,3), (3,1)]$ resolutions using the AnyRes techniques \cite{liu2024llavanext}. Additionally, our vision encoder remains frozen during training, which restricts its ability to learn and adapt. In future work, we plan to develop improved learning algorithms and models to enhance the model's visual understanding capabilities.

\subsubsection{CPT and IFT Dataset Ablations on MathVerse}

\begin{table}[t]
\begin{center}
\begin{minipage}{0.45\textwidth}
\caption{Datasets ablations (CPT and IFT) using Deepseek-coder-1.3B.}
\centering
\begin{tabular}{cccl}
\toprule
              & CPT         & IFT      & \begin{tabular}[c]{@{}l@{}}MathVerse\\ w/o score\end{tabular} \\
\midrule
\scriptsize{DSC-1.3B} &             & Mavis     & 20.2 \\
\scriptsize{DSC-1.3B} & \checkmark  & Mavis     & 25.1 \scriptsize{(+4.9)} \\
\scriptsize{DSC-1.3B} &             & Extended  & 22.3 \\
\scriptsize{DSC-1.3B} & \checkmark  & Extended  & 26.9 \scriptsize{(+4.6)} \\
\bottomrule
\end{tabular}
\label{table:alabtion-mathverse-1.3b}
\end{minipage}\hfill
\begin{minipage}{0.45\textwidth}
\centering
\caption{Datasets ablations (CPT and IFT) using Deepseek-coder-1.5-7B.}
\begin{tabular}{cccl}
\toprule
              & CPT         & IFT      & \begin{tabular}[c]{@{}l@{}}MathVerse\\ w/o score\end{tabular} \\
\midrule
\scriptsize{DSC-1.5-7B} &             & Mavis     & 22.8 \\
\scriptsize{DSC-1.5-7B} & \checkmark  & Mavis     & 27.1 \scriptsize{(+4.3)} \\
\scriptsize{DSC-1.5-7B} &             & Extended  & 23.8 \\
\scriptsize{DSC-1.5-7B} & \checkmark  & Extended  & 29.1 \scriptsize{(+5.3)} \\
\bottomrule
\end{tabular}
\label{table:alabtion-mathverse-7b}
\end{minipage}
\end{center}
\end{table}

In this section, we conduct ablation studies on models (1) trained with and without continue pre-training (CPT), and (2) models fine-tuned on the MAVIS dataset versus a more extensive instruction fine-tuning (IFT) dataset. Specifically, we compare models trained with and without our own mathematical multi-modal pre-training dataset, InfiMM-WebMath-40B. Additionally, we evaluate two IFT dataset configurations: (a) a combination of MAVIS-Caption-to-QA, MAVIS-Existing-Dataset-Augment, MAVIS-Caption, MAVIS-DataEngine-Geometry, and MAVIS-Meta-Question (referred to as the MAVIS dataset); and (b) a broader set consisting of the MAVIS datasets along with Vflan, VisualWebInstruct, AI2D, CHARTQA, DOCVQA, DVQA, GEOQA, DART-Math, and Numina-Math (referred to as the Extended dataset).

As shown in Table~\ref{table:alabtion-mathverse-1.3b}, in the 1.3B model, CPT improves the MathVerse scores by 4.9 and 4.6 points when IFT is performed with MAVIS and Extended datasets, respectively. Similarly, Table~\ref{table:alabtion-mathverse-7b} shows that in the 7B model, CPT improves the MathVerse scores by 4.8 and 5.3 points with MAVIS and Extended datasets, respectively. In contrast, using broader IFT datasets typically enhances model performance by approximately 2 points. These results highlight the significant mathematical capabilities imparted to the models through our InfiMM-WebMath-40B for CPT.

\subsubsection{Evaluations on We-Math}

\begin{table}[!ht]
\begin{center}
\caption{Evaluation of models on the We-Math benchmark. AVG represents the primary metric of interest.}

\begin{tabular}{ccccccc}
\toprule
Model                           & Base LLM                                  & \cellcolor{gray!30} AVG $\uparrow$  & IK $\downarrow$ & IG $\uparrow$ & CM $\uparrow$ & RM $\downarrow$   \\ \midrule
\multicolumn{7}{c}{\textit{Proprietary Models}}                                       \\ \midrule
Qwen-VL-Max                     & \scriptsize{N/A}                            & \cellcolor{gray!30} 10.5  & 65.1   &  7.6   &  6.7   & 75.5 \\
Gemini-1.5-Pro                  & \scriptsize{N/A}                            & \cellcolor{gray!30} 26.4  & 42.7   & 11.2   & 20.8   & 54.8 \\
GPT-4V                          & \scriptsize{N/A}                            & \cellcolor{gray!30} 31.1  & 39.8   & 14.5   & 23.8   & 47.9 \\
GPT-4o                          & \scriptsize{N/A}                            & \cellcolor{gray!30} 42.9  & 31.2   & 15.2   & 35.2  & 34.2 \\ \midrule
\multicolumn{7}{c}{\textit{Open-sourced Models}}                                     \\ \midrule
LLaVA-1.5                       & \scriptsize{Vicuna-7B}                    & \cellcolor{gray!30} 6.5   & -   & -   & -   & 85.6 \\
LLaVA-1.5                       & \scriptsize{Vicuna-13B}                   & \cellcolor{gray!30} 8.4   & -   & -   & -   & 78.1 \\
LLaVA-1.6                       & \scriptsize{Vicuna-7B}                    & \cellcolor{gray!30} 3.3   & 78.3   & 2.5   & 2.1   & 89.1 \\
LLaVA-1.6                       & \scriptsize{Vicuna-13B}                   & \cellcolor{gray!30} 5.2   & 69.1   & 3.2   & 3.6   & 86.9 \\
LLaVA-NeXT                      & \scriptsize{Mammoth2-7B}                  & \cellcolor{gray!30} 13.4  & -   & -   & -   & 71.0 \\
LLaVA-NeXT                      & \scriptsize{Qwen-1.5-110B}                & \cellcolor{gray!30} 19.2  & -   & -   & -   & 66.0 \\
DeepSeek-VL                     & \scriptsize{DeepSeek-7B}                  & \cellcolor{gray!30} 6.3   & 69.1   & 4.6   & 4.0   & 84.8 \\
G-LLaVA                         & \scriptsize{Vicuna-13B}                   & \cellcolor{gray!30} 6.5   & 64.2   & 4.6   & 4.2   & 86.6 \\
Math-LLaVA                      & \scriptsize{Vicuna-13B}                   & \cellcolor{gray!30} 11.1  & -   & -   & -   & 72.8 \\
InternLM-XC2                    & \scriptsize{InternLM2-7B}                 & \cellcolor{gray!30} 12.7  & 56.4   & 10.5   & 7.4   & 77.6 \\ \midrule
\multicolumn{7}{c}{\textit{Our Models}}                                                \\ \midrule
InfiMM-Math                     & \scriptsize{DeepSeek-Coder-1.3B}          & \cellcolor{gray!30} 13.1  & 56.2   & 9.1   & 9.3   & 73.7 \\
InfiMM-Math                     & \scriptsize{DeepSeek-Base-7B}             & \cellcolor{gray!30} 20.6  & 48.8   & 12.2  & 15.2  & 61.7 \\ \bottomrule
\end{tabular}
\label{table:eval-wemath}
\end{center}
\end{table}

In this section, we compare models evaluated on the We-Math benchmarks. We-Math consists of 6.5K visual math questions and employs a four-dimensional metric based on the different levels of knowledge required to answer each question: Insufficient Knowledge (IK), Inadequate Generalization (IG), Complete Mastery (CM), and Rote Memorization (RM). We report results on the We-Math \texttt{testmini} set using all four metrics.

As shown in Table \ref{table:eval-wemath}, our model, InfiMM-Math, surpasses all open-source models. 
Notably, our model with a 7B base LLM outperforms LLaVA models with 72B and 110B LLMs. When compared to math-specific models like LLaVA, Math-LLaVA, G-LLaVA and InternLM-XC2 of similar size, our model demonstrates significant improvements. For instance, compared to LLaVA-NeXT-7B and InternLM-XC2, we achieve an improvement of 7.2 and 7.9 points in AVG scores, respectively.
\section{Conclusions}
In this work, unlike most current works that focus on instruction-following datasets to enhance LLMs' mathematical reasoning, we introduce InfiMM-WebMath-40B, the first open source large-scale multimodal interleaved mathematical pretraining dataset, addressing a critical gap in the multimodal research community.
We provide detailed descriptions of our data collection process and publicly release the dataset to support open-source research. Our model, InfiMM-Math, demonstrates exceptional performance on the MathVerse and We-Math benchmarks, highlighting the effectiveness of the proposed dataset.

For future work, we aim to further advance mathematical reasoning in Multimodal Large Language Models (MLLMs) by exploring more sophisticated components specifically designed for mathematical content. 
This includes developing enhanced vision encoders tailored to effectively process mathematical symbols, diagrams, and equations. 
Additionally, we plan to integrate reinforcement learning techniques to improve reasoning capabilities in mathematical contexts. 
Beyond these efforts, we will explore other innovative approaches to push the boundaries of mathematical understanding in MLLMs, addressing the unique complexities inherent in multimodal mathematical reasoning.

\clearpage

{\small
\bibliographystyle{ieee_fullname}
\bibliography{egbib}
}

\newpage
\appendix
\section{Using Prompting with Llama-3-70B for Mathematical Annotation}
\label{sec:llama3:prompt}
\begin{table}[ht]
\centering
\caption{Prompt for evaluating mathematical content using Llama-3-70B following FineWeb-Edu~\cite{lozhkov2024fineweb-edu}.}
\label{tab:llama3:prompt}
\begin{minipage}{\textwidth} 
\tiny 
\ttfamily
\raggedright
Below is an extract from a web page. Evaluate the mathematical value of the extract and its potential utility as a teaching resource in a mathematical context using the additive 10-point scoring system described below. Points accumulate based on the satisfaction of each criterion, with special attention to the presence and quality of mathematical equations:

- 0 points if the extract includes no mathematical content, such as only provides historical context, summarizes an article's abstract, or exclusively features a person's resume.

- 1-2 points if the extract offers rudimentary information on mathematical subjects, even if interspersed with irrelevant material such as advertisements or non-academic content.

- 2-4 points if the extract touches upon mathematical topics without rigorous adherence to academic standards and contains a mix of mathematical and non-mathematical content, or if the presentation is haphazard and the writing lacks clarity.

- 4-6 points if the extract presents key concepts pertinent to educational curricula and includes mathematical equations, albeit potentially non-comprehensive or alongside superfluous information. It should resemble a mathematical text, such as an introductory section of a textbook or a basic tutorial.

- 6-8 points if the extract is highly relevant to mathematics, is well-structured, and offers a clear exposition, including a significant number of mathematical equations and solutions. It should be akin to an in-depth textbook chapter or tutorial, with a strong focus on mathematical content and minimal unrelated information.

- 8-10 points if the extract exhibits exceptional mathematical merit, characterized by detailed explanations, a comprehensive array of mathematical equations, and a coherent, accessible writing style that provides profound insights into mathematical theories and applications.

The extract:
<EXAMPLE>.

After examining the extract: 
- Briefly justify your total score.
- Conclude with the score using the format: "mathematical score:  <total points>"
\end{minipage}
\end{table}
\section{Ablation Studies on High-Precision Mathematical Content Filtering}
\label{sec:ablation:hp:filtering}
In this section, we examine the efficacy of two classifiers—LLM-based and fastText-based—focusing on high-precision mathematical content filtering. 
The comparison utilizes the DeepSeek-Coder 1.3B model, which we trained on a dataset previously introduced in Sec.~\ref{sec:initial_fasttext} with a sequence length of 4096. This model was trained to score documents based on their relevance to mathematical content on a scale from 0 to 10.

We conduct the continue pretraining of the DeepSeekCoder 1.3B model using datasets filtered by both the LLM- and fastText-based classifiers. 
Table~\ref{tab:ablation_on_filter} shows the performance results. 
The results highlight a length bias in the LLM-based method, which tends to favor longer documents, averaging 2,500 tokens, compared to 1,700 tokens for the FastText filter. 
The length bias associated with the LLM-based classifier has adversely impacted the dataset's performance on the GSM8K dataset. 
As indicated in the table, the LLM-filtered dataset achieved lower accuracy (17.5\%) on the GSM8K dataset compared to the fastText-filtered dataset (20.2\%). 
This decrease in performance indicates that the LLM's preference for longer documents may not align well with the requirements of datasets like GSM8K, which demand concise and precise mathematical descriptions.

Given these insights, we have decided to continue utilizing the fastText classifier for high-precision filtering in our ongoing research. 
Nonetheless, the implications of the LLM-based classifier require further investigation to fully understand and address its biases.

\begin{table}[ht]
\centering
\caption{Ablations on the high-precision filtering. The ``Text Avg Length'' column indicates the averaged document length after filtering by each respective classifier.}
\label{tab:ablation_on_filter}
\begin{tabular}{cccc}
\toprule
              & MMLU (STEM)        & GSM8K      & Text Avg Length\\
\midrule
LLM-Classifier &   32.8  & 17.5\%  & 2500 \\
FastText-Classifier &  31.1 & 20.2\%  & 1700 \\
\bottomrule
\end{tabular}
\end{table}

\end{document}